\documentclass[10pt,twocolumn,letterpaper]{article}

\usepackage[
  top=0.75in,
  bottom=1in,
  left=0.75in,
  right=0.75in,
  columnsep=0.3in
]{geometry}

\usepackage{times}
\usepackage[T1]{fontenc}
\usepackage{microtype}

\usepackage{amsmath,amssymb}

\usepackage{booktabs}
\usepackage{tabularx}
\usepackage{array}
\usepackage{multirow}

\usepackage{graphicx}
\usepackage[dvipsnames]{xcolor}
\usepackage[table]{xcolor}

\usepackage{enumitem}
\setlist{nosep,leftmargin=1.2em}

\usepackage[font=small,labelfont=bf,skip=4pt]{caption}

\usepackage{abstract}

\usepackage{titlesec}
\titleformat{\section}
  {\large\bfseries}{\thesection}{1em}{}
\titleformat{\subsection}
  {\normalsize\bfseries}{\thesubsection}{1em}{}
\titleformat{\subsubsection}
  {\normalsize\itshape}{\thesubsubsection}{1em}{}
\titlespacing*{\section}{0pt}{12pt plus 2pt minus 2pt}{4pt plus 1pt minus 1pt}
\titlespacing*{\subsection}{0pt}{8pt plus 2pt minus 1pt}{2pt plus 1pt minus 1pt}
\titlespacing*{\subsubsection}{0pt}{6pt plus 1pt minus 1pt}{2pt plus 1pt minus 1pt}

\usepackage{fancyhdr}
\usepackage[
  colorlinks=true,
  linkcolor=MidnightBlue,
  citecolor=MidnightBlue,
  urlcolor=MidnightBlue,
  bookmarks=true
]{hyperref}
\usepackage{cleveref}

\usepackage{authblk}

\usepackage[bottom]{footmisc}

\begin{document}

\twocolumn[
\begin{@twocolumnfalse}
\begin{center}
{\LARGE\bfseries Distilling Expert Judgment at Scale
 \par}
\vspace{12pt}
{\normalsize
Robbie Goldfarb,$^{1}$
Andrew B.\ Hall,$^{1,2}$
Jillian Fisher,$^{1,3}$
Kathryn Salam,$^{1}$
Matt Wilde$^{1}$
\par}
\vspace{6pt}
{\small\itshape
$^{1}$Forum AI \quad
$^{2}$Stanford University
$^{3}$University of Washington
\par}
\vspace{16pt}
\end{center}

\begin{onecolabstract}
Frontier AI models are increasingly deployed on topics where the stakes are high and expert judgment is required, including healthcare, defense, finance, governance, and politics. We present a general methodology for distilling expert human judgment into automated evaluation systems that operate at scale. Our approach assembles networks of domain experts and develops techniques for extracting not just their conclusions but their reasoning processes, encoding those processes into structured editorial standards that automated judges can apply continuously. We ground our approach in a long tradition of research on the nature of expertise and the conditions under which expert judgment is reliable and more effective than other approaches, and argue that independent, expert-calibrated evaluation is necessary not only on epistemic grounds but as a matter of accountability and public trust. We validate the methodology on AI news reporting as a first application domain. On source quality, two independent domain experts achieve Krippendorff's $\alpha = 0.83/0.69$ on a held-out test set when applying the editorial standards, and our automated judge reaches $F1 = 0.81/0.97$ against expert consensus. On neutrality, experts achieve $\alpha = .72$ on the test set, and the automated judge reaches $F1 = 0.86$ against expert consensus. Additionally, a controlled comparison shows that expert-calibrated judges substantially outperform uncalibrated frontier models on every source-quality metric, demonstrating that encoding expert reasoning is the critical step that prompt engineering alone cannot replace. We present this benchmark as a proof of concept for a broader evaluation infrastructure applicable to any domain where judgment matters and expertise is scarce.
\end{onecolabstract}
\vspace{16pt}
\end{@twocolumnfalse}
]

\section{Introduction}

Frontier AI models have gotten remarkably good at a wide range of tasks. But on the topics where the stakes are highest, the critical question is whether the model exercises sound judgment. Healthcare assistants must demonstrate clinical nuance and sensitivity to error; defense-related systems require careful escalation control; financial advising involves reasoning under uncertainty and risk-aware judgment; informational tools must maintain principled neutrality when faced with contested viewpoints. On these questions, there is currently no credible, independent infrastructure for finding out how well AI systems perform or helping them do better.

AI companies have published increasingly sophisticated self-evaluations, academic researchers have developed clever measurement instruments, and governments are beginning to mandate evaluation requirements, but each of these approaches faces structural limitations. Companies cannot credibly grade their own homework. Academic benchmarks tend to measure narrow constructs like political compass scores and refusal rates rather than the holistic epistemic quality that determines whether a model's output is actually trustworthy. And regulators, while they can mandate evaluation, rarely possess the deep domain expertise needed to design and validate the evaluation criteria themselves.

Our core idea is to fill this gap by distilling human expertise into automated evaluation systems that can operate at scale. We assemble networks of domain experts, professionals who have spent careers learning to reason carefully when the evidence is incomplete and the stakes are high, and develop techniques for extracting not just their conclusions but their reasoning processes. How do they weigh competing evidence? Where do they draw the line between legitimate disagreement and misleading framing? What distinguishes a genuinely truth-seeking response from one that merely avoids obvious offense? We aim to encode those reasoning processes into structured editorial standards that automated AI judges can apply continuously, combining the epistemic rigor of expert human judgment with the speed and coverage of automated systems.

The result is an evaluation infrastructure that produces assessments that are credibly independent, empirically validated against expert judgment, and capable of operating at the speed of breaking news. The framework is deliberately flexible, designed to evolve alongside the evaluation methods it implements. It applies to simple one-shot evaluations and extends naturally to the multi-turn and open-world settings that matter more as model capabilities advance.

The purpose of this whitepaper is to explain why we do this, how we do this, and to offer initial evidence of our approach's validity. First, we ground our approach in a long research tradition on the nature and limits of expertise, making an explicit case for why expert judgment is epistemically irreplaceable and why independent evaluation is structurally necessary (Section~\ref{sec:experts}). Second, we present the general methodology for turning expert reasoning into calibrated, automated evaluation (Sections~\ref{sec:method}--\ref{sec:validation}). Third, we validate the methodology on AI news reporting as a first application domain (Section~\ref{sec:application}), focusing on citation source quality and response neutrality, and demonstrating that expert-calibrated AI judges match expert human consensus and substantially outperform uncalibrated approaches. However, the architecture is designed to generalize to any domain where judgment matters and expertise is scarce, and building out that broader infrastructure is the work ahead.

\section{Why Expert Judgment Is Irreplaceable}
\label{sec:experts}

A natural response to the gaps we just described is to use more data, more crowdsourced raters, or more capable AI models. But we argue that this misses something fundamental about the nature of expertise. A long research tradition in cognitive psychology and decision science identifies conditions under which expertise is truly irreplaceable, and those conditions map precisely onto the domains where AI evaluation matters most.

\subsection{When Expert Judgment Is Necessary}

In a classic study, Kahneman and Klein (2009) established that expert intuition is valid when the environment contains sufficient regularity to be learnable and the expert has had adequate opportunity to learn those regularities through prolonged practice with feedback. Healthcare diagnosis, intelligence analysis, legal reasoning, financial risk assessment, and the evaluation of sensitive content are arguably all domains where important tasks may meet these conditions; they contain genuine regularities that professionals spend years learning to recognize, and they provide the kind of feedback that allows expertise to develop and be maintained.

Shanteau (1992) reached similar conclusions, arguing that expert competence is strongest in tasks with decomposable problems, verifiable standards, and regular feedback. Similarly, it has also been shown that expertise tends to break down in domains lacking these properties, such as long-range political prediction (Tetlock, 2005). At the same time, Tetlock’s later work offers a more constructive lesson: when expert judgment is paired with structured methodologies and regular calibration, ``superforecasters'' significantly outperform both unstructured experts and purely statistical approaches (Tetlock and Gardner, 2015). Our methodology is motivated by this perspective, seeking to structure and calibrate expert knowledge rather than relying on unstructured intuition alone.

The cognitive science of expertise clarifies why expert judgment is qualitatively different from non-expert judgment, not merely more of the same. Experts represent problems in terms of deep structural features rather than surface characteristics (Chi, Glaser, and Farr, 1988; Ericsson, Krampe, and Tesch-R\"{o}mer, 1993; Dreyfus and Dreyfus, 1986). A novice evaluating an AI response about a medical condition might focus on tone or length; a clinician will assess whether the response correctly triages urgency, distinguishes established protocols from contested approaches, and calibrates its confidence to the available evidence.

A deeper lesson, related to measurement instruments, comes from the clinical vs. statistical prediction literature. Meehl (1954) and subsequent meta-analyses (Dawes, Faust, and Meehl, 1989; Grove et al., 2000) showed that simple mechanical rules often outperform \emph{unstructured} clinical judgment. But these findings do not imply that complex evaluation can be reduced to naive scalar metrics or simplistic benchmarks. Rather, they suggest that expertise becomes most reliable when translated into structured, explicit procedures. Our methodology is motivated by this perspective, aiming to translate expert evaluation criteria into formalized procedures that remain faithful to the underlying judgment while remaining scalable. 

\subsection{Current Frameworks Often Lack Expertise}

In the absence of expert guidance, current evaluation frameworks typically rely on alternative forms of scalable supervision. Broadly, three main approaches have emerged for assessing AI outputs at scale.

\textbf{Crowdsourced evaluation} works well for tasks accessible to non-specialists, like simple image labeling or sentiment classification. It fails for tasks that require domain knowledge. Whether an AI system's analysis of a geopolitical crisis is sound or its health advice is clinically appropriate are questions that non-specialists cannot reliably answer, because they lack the structured domain knowledge that evaluation requires.

\textbf{LLM-as-judge approaches} use frontier AI models to evaluate other AI models. But inter-model agreement does not imply alignment with expert judgment. Two language models may agree that a response is ``balanced,'' but neither may agree with what a domain expert would conclude. Our empirical results (Section~\ref{sec:application}) demonstrate this concretely: an uncalibrated frontier model achieves dramatically lower agreement with expert consensus on every metric, and iterative prompt engineering does not close the gap. The missing ingredient is the encoding of expert reasoning that prompt engineering alone cannot reproduce.

\textbf{Simplified measurement instruments} like opinion surveys and refusal-rate benchmarks measure narrow constructs that miss holistic epistemic quality. A model can score well on a political compass test while routinely citing unreliable sources, presenting false balance on settled questions, or hallucinating claims about breaking news.

\subsection{The Case for Independent Evaluation}

Beyond the epistemic argument, there is a structural case for expert-grounded evaluation that improvements in AI capability alone cannot resolve.

AI systems deployed on consequential topics affect people's lives in ways that demand external accountability. The entities building these systems cannot credibly evaluate their own products, for the same structural reasons that companies cannot audit their own financial statements or that pharmaceutical firms cannot approve their own drugs. Credible evaluation requires visible independence from the entity being evaluated.

Several major AI companies have published self-evaluations of how their models handle sensitive content. These efforts are often methodologically sophisticated and commendably transparent. But they face an irreducible structural limitation in that the evaluator and the evaluated are the same entity. On the other hand, government oversight faces complementary limitations as the regulators can mandate evaluation but rarely possess the domain expertise needed to design criteria and validate judges. The result is an institutional gap that our methodology is designed to fill, combining the domain expertise of independent professionals with the scalability of automated systems.

\section{How We Turn Expertise into Evaluation}
\label{sec:method}

We now describe the general methodology we have developed for distilling human expertise into automated evaluation systems. 

\subsection{Expert Selection and Panel Design}

The foundation of our methodology is a network of domain experts who define what good evaluation looks like and calibrate our automated judges to apply those standards as expert. Once calibrated, the LLM judges run independently and can rapidly provide assessments of new content.

Selecting the right experts matters enormously. Domain knowledge is necessary but not sufficient. What we look for is \emph{evaluative judgment}, which is the ability to assess reasoning under uncertainty, separate evidence from interpretation, and identify when a superficially adequate response is smuggling in assumptions. We favor people whose professional work involves having their conclusions challenged. Analysts, investigators, scholars, and practitioners who have spent careers refining how they think about hard questions, not just what they think about them. Before joining a panel, every candidate completes structured calibration exercises on pre-scored and ambiguous cases, designed not to check whether their answers match ours but to verify the thoroughness and discipline of their reasoning process.

Additionally, cross-perspective credibility is a structural requirement as well. If the expert layer that calibrates the system is perceived as representing one viewpoint or constituency, the entire enterprise fails on its own terms. For each domain, we assemble panels that span the relevant axes of professional and intellectual diversity so that the evaluation criteria encode shared analytical standards rather than any faction's preferences.

\subsection{Extracting Reasoning Processes, Not Just Conclusions}

Before constructing test sets or calibrating AI judges, we first need to understand what ``good'' actually looks like within a given domain. To do this, we conduct structured interviews with our expert network to map the broader evaluative landscape. We note that the experts involved in this stage are senior practitioners with extensive real-world decision-making experience in their respective domains, including former intelligence directors, hospital leaders, and executives of major financial institutions. This process identifies not only which topics are sensitive, but also where informed professionals genuinely disagree in good faith and what distinguishes an epistemically responsible response from one that merely avoids obvious errors. These interviews form the foundation for multiple stages of our evaluation development pipeline, including identifying which dimensions of a domain should be evaluated, selecting representative and high-stakes test prompts, and constructing the standards used for evaluation.

The challenge is extracting generalizable evaluation standards from expert input without encoding any individual expert’s personal views into the system. Over several hundred hours of expert interviews across domains, we developed a structured elicitation process adapted from cognitive task analysis methods (Klein, 1998). Rather than asking experts for abstract principles alone, we ground discussions in concrete AI-generated artifacts and probe the cues, comparisons, and counterfactuals underlying their judgments. We then compare reasoning patterns across experts and scenarios to identify recurring standards and shared failure modes without privileging any single perspective. The goal throughout is to transform tacit expert judgment into explicit, reproducible evaluation criteria that automated systems can apply consistently at scale.

\subsection{Building Test Sets That Mirror Real Users and Probe Real Weaknesses}

The prompts used to evaluate models must balance two goals in genuine tension. The test set has to mirror how real users actually interact with AI systems on a given topic, with all the messy phrasing, unclear intent, and varying complexity that entails, because a set built entirely from polished expert questions will miss important failure modes. But it also has to systematically probe known weaknesses, cover the full evaluation grid, and include enough high-risk prompts to produce statistically meaningful findings.

We characterize every prompt along multiple coverage dimensions, including the substantive domain, the task type, and the cognitive complexity. We draw prompts from three sources. \emph{Real-world prompts} are sourced from places where people actually ask questions on the topic, preserved with minimal editorial cleanup. \emph{Expert-generated prompts} are crafted by evaluators to probe scenarios that organic queries rarely cover. \emph{Synthetic prompts} fill coverage gaps and generate companion prompts for controlled comparisons.

A fixed portion of the test set (targeting 65\%) remains stable across evaluation cycles to enable longitudinal comparison. The remaining portion is refreshed as the domain evolves. The core set is locked before each evaluation run and does not change in response to model performance. We publish our standards and validation methodology but do not publish our test sets, because disclosure would allow vendors to optimize for our prompts rather than improving general capability.

\subsection{From Expert Interviews to Automated Judges}

After conducting extensive expert interviews and constructing test sets, the core technical challenge becomes building automated AI judges that can evaluate model responses at scale with the rigor of expert reviewers. This process involves three stages. First, synthesizing expert reasoning into a draft set of editorial standards. Second, operationalizing those standards into a labeling protocol that robustly produces calibration data. Third, engineering an automated judge that reliably reproduces expert labels.

\subsubsection{From Expert Interviews to Editorial Standards}

The process begins with structured interviews with senior domain experts who evaluate the same model outputs and explain their reasoning in detail. Across these sessions, a recurring pattern emerges. Experts from different professional backgrounds, while differing in emphasis and ordering, tend to follow a broadly similar sequence of evaluative steps. They first determine what kind of response the input warrants, then assess how the model handles the input’s premises, examine the output for domain-specific quality markers, and finally evaluate the overall epistemic adequacy of the response. Experts may disagree on difficult edge cases, but the underlying analytical structure is remarkably consistent.

Using these interviews, a small group of internal experts synthesizes the transcripts into an initial set of editorial standards, mapping recurring evaluative patterns to specific criteria. Each standard is operationalized through a categorical or ordinal rating scale paired with a free-response rationale field, allowing raters to explain the basis for their judgments. The draft standards are then iteratively refined against concrete examples until they can be applied consistently across a diverse set of external experts. The central design constraint is that the standards capture the structure of expert reasoning, rather than merely the conclusions experts happen to reach, translating that reasoning into procedures that can be applied reliably across cases and raters. This approach enables us to leverage not only shared reasoning in clear consensus cases, but also, and often more importantly, in ambiguous cases where expert disagreement or uncertainty is most informative. 

\subsubsection{From Standards to Golden Sets}

Once the editorial standards have been operationalized into a reliable labeling protocol, we construct a gold-labeled dataset for AI judge calibration. A pool of expert domain raters, largely distinct from the senior experts who developed the standards, applies the protocol to a diverse set of model responses. These raters are selected based on demonstrated domain expertise and sustained professional experience, such as tenured scholars whose research directly addresses the evaluation domain or journalists who have covered a specific beat for decades. Their role is not to redefine the standards, but to apply them consistently across cases. This process produces an expert-labeled gold set that serves as the reference standard for calibrating and evaluating automated judges.

We measure inter-rater reliability on the golden set using Krippendorff's $\alpha$, which accounts for agreement that could occur by chance and supports both binary and multi-category labels. When disagreements surface during labeling, they serve a dual purpose, first a senior expert adjudicates the individual item to produce the consensus label, but also the pattern of disagreements feeds back into the labeling protocol itself. Persistent disagreement on a standard often signals ambiguity in the definition or a rating scale that does not divide the space at the right joints, at which point we revise and iterate.

We also allow for disagreements among experts that do not reflect task ambiguity but, rather, reflect what we call ``reasonable pluralism." Sometimes, reasonable experts simply disagree at a deep level about something. In these cases, we escalate the disagreement, asking a larger group of experts to make the decision following the same guidelines. If this group cannot agree, the final decision is made via majority vote. 

\subsubsection{From Golden Sets to Automated Judges}

With the editorial standards and a golden set in hand, the engineering task is building an automated AI judge that reproduces those human labels. Each judge is built to be as simple as the evaluation task allows. A judge that classifies a single attribute from a short input can be a single LLM call; a judge that must extract dozens of claims, triage them, and independently verify each one against external evidence necessarily requires more machinery. There is no fixed architecture. Each judge uses the minimum structure needed to produce expert-level agreement on its specific evaluation dimension.

Where the judges share a design philosophy is in how they handle intermediate results. Data flows between steps as typed, structured objects rather than natural language summaries that a downstream LLM must interpret. This eliminates a common failure mode in LLM pipelines where an orchestrating model silently drops, summarizes, or distorts intermediate results. Where a step requires no judgment (aggregating scores, applying deduction rules, enforcing constraints), it is implemented in deterministic code rather than delegated to a model. LLM calls are reserved for the steps that genuinely require judgment, and everything else is handled by code that can be verified and audited.

AI judge calibration is evaluated against the gold set using the same Krippendorff's $\alpha$ metric employed during the human-rater stage, allowing human--human and human--judge reliability to be compared on a common scale. Calibration proceeds through an AI-assisted iterative refinement loop that targets the specific standards with the lowest agreement scores. After each modification, the judge is re-evaluated on the gold set, and every iteration is manually reviewed by a human evaluator who inspects the AI-generated execution trace before interpreting the resulting metrics.

Importantly, once the human labeling protocol is finalized in the previous stage, the editorial standards themselves remain fixed. The calibration process is therefore one-directional, refining the AI judge to better reproduce expert labels rather than altering the meaning of the labels to fit the judge. Iteration continues until either agreement no longer improves or the AI judge reaches a reliability level comparable to human--human agreement. 

\section{Two Tests for Trustworthy Evaluation}
\label{sec:validation}

Our evaluation pipeline produces large volumes of judgments through automated judges. To ensure these judgments remain faithful to expert reasoning, we validate the system at two levels. First, we test whether experts can apply the editorial standards consistently. Second, we test whether the automated judge reproduces the decisions those experts would make. Together, these two tests assess both the reliability of the evaluation framework itself and the fidelity of its automation.

\paragraph{Test 1: Can experts apply the standards consistently?}

Before an automated AI judge can be trusted, the underlying editorial standards must be shown to support reliable human expert application. We therefore recruit multiple domain experts to independently evaluate the same set of model responses using the editorial standards and labeling protocol, without visibility into one another's judgments. To ensure reliable estimation across all rating categories, we oversample classes that are rare in the standard prompt distribution.

We measure inter-rater reliability using Krippendorff's $\alpha$ (Krippendorff, 2019; Hayes and Krippendorff, 2007). Following Krippendorff's guidance for moderate agreement, we target a threshold of $\alpha \geq 0.67$ (Krippendorff, 2019). When agreement falls below this threshold, senior experts review disagreement patterns to identify ambiguities in the standards or labeling protocol, which are then iteratively refined. This test provides evidence that the standards have been operationalized clearly enough to be applied consistently across a diverse set of qualified experts.

\paragraph{Test 2: Can the automated judge reproduce expert decisions?}

Once the editorial standards and expert-labeled gold set are finalized, we evaluate whether the automated AI judge reproduces the same decisions. The judge evaluates the same responses, and we compare its labels against adjudicated expert labels using precision, recall, and F1 on a held-out test set. This allows us to directly measure how well AI judges can reliably reproduce expert judgment.

We use $F1>0.70$ as the target threshold as it indicates moderate agreement for most everyday tasks, treating performance at or above this level as evidence that the automated AI judge has been successfully calibrated to expert judgment. Together, these two validation layers ensure that experts apply the rubric consistently and that the automated judge reliably reproduces those decisions, enabling expert evaluation to scale while preserving methodological rigor.

\section{Application: AI News Reporting}
\label{sec:application}

We validate our methodology on the problem of evaluating how frontier AI models handle reporting of news content. News content is a natural first domain for several reasons. It is among the most visible applications of AI to the general public, the consequences of poor judgment are immediately apparent and politically salient, and AI companies have already attempted self-evaluation. These properties make news reporting not only a natural starting point, but also a domain that demands careful, credible, and domain-specific evaluation design. 

\subsection{Domain-Specific Design}

Our expert network for this domain includes professionals drawn from intelligence analysis, foreign policy, journalism, law, economics, and academia. Our bipartisan advisory board includes figures ranging from distinguished former politicians and senior policymakers to leading journalists and academic researchers. Cross-ideological credibility is essential. If the expert layer is perceived as representing one political camp, the enterprise fails on its own terms.

The expert elicitation process described in Section~\ref{sec:method} produced editorial standards organized around three dimensions. \emph{Source quality} asks whether cited sources are authoritative and appropriate. \emph{Factual accuracy} asks whether verifiable claims in the response are correct. \emph{Neutrality} asks whether information is presented without framing that implicitly favors particular political positions. For this dimension, we evaluate not only the response but also the input for context. For example, an input which requests the model argue a specific position is not penalized for advancing one side.

We have prioritized validation work on two of these dimensions, source quality and neutrality, and present results for both in this paper. Factual accuracy involves additional pipeline machinery (claim extraction followed by claim verification) and is at an earlier stage of validation; we report on it in subsequent work. The methodology presented here is designed to extend naturally to additional editorial standards within the political domain and to other application domains over time.

\begin{table*}[t]
\centering
\caption{Coverage dimensions for prompt characterization.}
\label{tab:coverage}
\small
\begin{tabularx}{\textwidth}{@{}l p{3.8cm} X@{}}
\toprule
\textbf{Dimension} & \textbf{Description} & \textbf{Categories in Our Benchmark} \\
\midrule
Domains & Substantive policy area, ensuring breadth across major political and geopolitical issue spaces. &
Government \& Politics; Economic Policy; Social Policy; Environmental \& Energy; Technology \& Society; Foreign Policy \& Security; Transnational Issues; Conflict and War; Global Economy; + customer- or product-specific tasks \\
\addlinespace
Task types & Functional form of the user request, capturing \emph{how} the model is being asked to respond. &
Informational queries; Perspective role-play; Policy deliverables; Charged framings; + customer- or product-specific tasks \\
\addlinespace
Complexity & Cognitive and normative difficulty, from objective fact recall to value-laden reasoning. &
Factual; Interpretive; Analytical; Normative; Strategic \\
\bottomrule
\end{tabularx}
\end{table*}

We draw prompts from three sources: real-world prompts from forums and social media where people actually ask politically sensitive questions; expert-generated prompts designed to probe adversarial framings, edge cases, and specific bias types; and synthetic prompts that fill coverage gaps and generate ideological companion prompts for controlled comparisons. Each prompt is also characterized by temporal context (historical events and current debates), because models are most likely to hallucinate when facts are still in flux.

Table~\ref{tab:coverage} summarizes the three coverage dimensions along which we characterize every prompt.

\subsection{Source Quality}
Our source quality judge evaluates each citation in an LLM response using two standards defined by an expert panel. First, it assigns every source to one of six categories, indicating the type of source the citation comes from: primary, scholarly, think tank, journalistic, commercial content, or informal. Second, it applies a binary flag to indicate whether the outlet is state-controlled, meaning its editorial independence is compromised (e.g., RT, CGTN, Xinhua). State-funded but editorially independent broadcasters (e.g., BBC, NPR, PBS, Deutsche Welle) are not flagged. These two signals are then combined through a formula developed by the same expert panel to produce an overall assessment of source quality for the response on a 0--100 scale. 

\subsubsection{Human--Human Agreement}
To evaluate the robustness of applying the editorial standards among domain experts, we have two domain experts who did not participate in standard development independently labeled sources using the final editorial standards, with a senior expert adjudicating any disagreements. We report agreement separately on the validation pair (used for AI judge calibration; $n \approx 230$) and on a held-out test set ($n \approx 150$). Inter-rater agreement is measured using Krippendorff’s $\alpha$, with $95\%$ confidence intervals estimated via item-level bootstrapping ($2K$ resamples). 

\begin{table}[h]
\centering
\caption{Source Quality: Inter-expert agreement (Human--Human). Krippendorff's $\alpha$ with item-bootstrap 95\% CI on both the validation and test set.}
\label{tab:sq-human}
\small
\begin{tabular}{@{}lcc@{}}
\toprule
\textbf{Standard} & \textbf{Validation } & \textbf{Test $\alpha$} \\
& \textbf{$\alpha$ [95\% CI]}& \textbf{$\alpha$ [95\% CI]}\\
\midrule
Source type (categ.) & .82 [.76, .87] & .83 [.76, .89] \\
State-controlled (bin.) & .93 [.80, 1.00] & .69 [.52, .83] \\
\bottomrule
\end{tabular}
\end{table}

\Cref{tab:sq-human} shows that, across both standards, the two expert evaluators achieve high to moderate inter-rater reliability. The categorical source-type standard is high ($\alpha = .83$ on test) and essentially identical between the validation and test sets, showing that the editorial standards generalize well to new experts. The state-controlled flag is rare in the test set, which inflates the chance-corrected sampling error and explains the wider confidence interval. Even so, there is moderate agreement ($\alpha = 0.69$ on test), exceeding our minimum acceptable threshold of $0.67$. Overall, this observed level of agreement indicates that the application of the editorial standards generalizes beyond its original authors and can be consistently applied by independent human experts.

\subsubsection{Judge--Human Agreement}
\label{sec:sq-judge-human}

Next, we evaluate how well the calibrated automatic AI judge can mimic the human expert annotations. To test this, the automated judge evaluated the same items in the held-out test set. We measure reliability using precision, recall, and F1 of the AI annotations against the human gold-labels.

\begin{table}[h]
\centering
\caption{Source Quality: Precision, recall, and F1 of the automated AI judge against the adjudicated human consensus on the test set. For Source, \emph{macro avg} is the unweighted mean across the six tiers; \emph{weighted avg} is support-weighted.}
\label{tab:sq-judge-class}
\small
\begin{tabular}{@{}lrccc@{}}
\toprule
\textbf{Standard} & \textbf{$n$} & \textbf{Prec.} & \textbf{Rec.} & \textbf{F1} \\
\midrule
\multicolumn{5}{@{}l}{Source type (categ.)} \\
\quad \textit{macro avg} & 146 & .74 & .80 & .75 \\
\quad \textit{weighted avg} & 146 & .86 & .79 & .81 \\
\midrule
\multicolumn{5}{@{}l}{State-controlled (bin.)} \\
\quad \textit{macro avg} & 147 & .98 & .92 & .95 \\
\quad \textit{weighted avg} & 147 & .97 & .97 & .97 \\
\bottomrule
\end{tabular}
\end{table}

In \Cref{tab:sq-judge-class}, it can be seen that the judge reaches $F1 = .81/.97$ on the six-class source-type standard and binary state-controlled standard, which both reflect a high degree of reliability of the AI judge compared to human expert judgment. We note that, in general, both precision and recall are high which indicates balanced performance on all classes. This is especially important for Source type, which has six different classes. These results show that the calibrated AI judge successfully reproduces expert application of the editorial standards on the source-quality task and can be used for scaling expert judgment.

\subsection{A Capable Model Alone Cannot Close the Gap}
\label{sec:sq-baseline}
To further evaluate whether expert guidance provides measurable benefits over non-expert approaches, we conduct a preliminary comparison between our final expert-guided AI judge (using Claude Code) and two non-expert-guided alternatives. Specifically, we provide an AI agent with a minimal task description for evaluating source quality and ask it to generate a rubric that produces a continuous source quality judgment. The agent is given access to example prompts, responses, and citations, but no gold-standard labels. The implementation process consists of an initial build phase followed by a single refinement iteration, during which the agent executes its implementation, inspects the resulting outputs, and revises accordingly. We run this experiment using two high-performing AI coding agents, Codex and Claude Code. Both produced rubrics which gave a (different) 1--5 numeric scales of source quality. 

Because the AI agents developed a different taxonomy than our expert-guided method, a direct
per-standard comparison is not possible. Instead, we measure
each system’s agreement with expert consensus on three
dimensions, computed over the same test set used in the previous analysis:
\begin{itemize}
    \item \textbf{Source ordering} (Spearman $\rho$): do the systems rank
sources the same way experts do?
    \item \textbf{Bad-source detection} (Cohen’s $\kappa$): do the systems identify the same problematic sources—defined as commercial content, informal, state-controlled, or out-of-core competency?
    \item \textbf{Response-level agreement} (Spearman $\rho$): do the systems correlate with experts on which \textit{overall} model responses have strong or weak sourcing?
    \item \textbf{Response-level error} (MAE): how closely do the systems match experts' \textit{overall} sourcing quality scores, measured as absolute deviation from expert ratings?\end{itemize}
We note that both response-level metrics (Spearman $\rho$ and MAE) are computed using the average score across all citations within a response.
\begin{table}[h]
\centering
\caption{Expert-guided AI judges vs.\ uncalibrated AI judges. Mean metric with bootstrap 95\% confidence intervals in brackets with $2K$ samples. \textbf{Best} value is bolded; lower is better for MAE.}
\label{tab:expert_nonexpert_results}
\small
\setlength{\tabcolsep}{2pt}
\begin{tabular}{l>{\columncolor{gray!15}}c c c}
\toprule
\textbf{Layer} & \textbf{Claude+Expert} & \textbf{Claude} & \textbf{Codex} \\
\midrule
\multicolumn{4}{l}{\textit{Agreement metrics (higher better)}} \\

Source ord. ($\rho$) &
\textbf{.76} [.67,.85] &
.27 [.14,.39] &
.59 [.49,.67] \\

Bad-source ($\kappa$) &
\textbf{.79} [.70,.86] &
.03 [-.09,.14] &
.30 [.17,.42] \\

Resp. agr. ($\rho$) &
\textbf{.37} [.26,.48] &
.14 [.02,.26] &
.16 [.04,.27] \\

\midrule
\multicolumn{4}{l}{\textit{Error metric (lower better)}} \\

Resp. err. (MAE) &
\textbf{12.8} [11.4,14.2] &
28.9 [26.7,31.1] &
27.6 [25.5,29.8] \\

\bottomrule
\end{tabular}
\end{table}

\paragraph{Results.}
Table~\ref{tab:expert_nonexpert_results} shows that expert guidance substantially improves agreement with expert consensus across all evaluation layers. The expert-guided judge achieves the strongest performance on source ordering ($\rho=.76$), substantially outperforming both Codex ($\rho=.59$) and Claude ($\rho=.27$). The largest gap emerges in bad-source detection, where the expert-guided system reaches near-substantial agreement with experts ($\kappa=.79$), while Codex shows only modest agreement ($\kappa=.30$) and Claude performs near chance ($\kappa=.03$). Gains also persist at the response level, where the expert-guided system more closely matches expert judgments of overall sourcing quality ($\rho=.37$ and MAE = $12.8$) than either baseline.

These findings suggest that capability alone is insufficient to recover expert evaluation standards. Even strong coding agents, given examples and an opportunity to iteratively refine their implementation, fail to match the consistency of an expert-guided approach. The gap is particularly pronounced for identifying problematic sources, suggesting that expert judgment encodes tacit distinctions that are difficult to infer from task descriptions and examples alone. Better understanding the contribution of expert input remains an important direction for future work, which we plan to investigate through targeted ablations. 

\subsection{Neutrality}
Neutrality is inherently context-dependent and does not admit a single scalar definition. Rather than attempting to capture it directly, we used structured interviews with senior experts to develop a set of editorial standards that jointly serve as reliable indicators of neutrality. Through this process, experts converged on a flowchart-based decision procedure that first classifies a prompt by type and then routes it to type-specific standards, reflecting the insight that neutrality is a property of the prompt--response pair rather than the response alone. 

The flowchart terminates in a binary verdict of \emph{approximate neutrality}, acknowledging that no fixed set of rules can fully resolve the underlying conceptual ambiguity. For responses that fail to meet these standards, we perform a secondary analysis to characterize the direction of any partisan lean. Currently, this analysis is limited to U.S. political orientations (left or right), though we plan to extend it to additional dimensions. This secondary evaluation was developed using the same expert-guided pipeline and similarly takes the form of a flowchart conditioned on prompt type.

The prompt taxonomy distinguishes several categories such as purely factual questions, directed prompts (where the user explicitly asks the model to argue a position), loaded prompts (containing false premises or pejorative framing), normative ``should'' questions, and open-ended questions that may use nuance framing. Each category carries its own neutrality standard, reflecting that balanced behavior looks different depending on prompt type.

For each prompt, neutrality is operationalized through a set of editorial standards composed of named rules, each paired with concrete pass/fail examples and explicit substitution or removal tests to improve reproducibility. These standards were crafted by senior experts and closely mirror editorial norms found in government agencies, political institutions, and major news and journalism outlets. Key principles include: faithfully executing a directed prompt is itself neutral, refusing to engage with a requested perspective is a neutrality failure; loaded premises must be corrected without overcorrecting; normative questions require balanced presentation without steering; and ``nuance'' framing must be substantiated by genuinely complex content rather than used as rhetorical cover for a one-sided response. The framework also addresses subtler failure modes, such as asymmetric attribution framing and false equivalence between evidence-strong and evidence-weak claims.

We make no claim that these are the only or best definitions of neutrality. Reasonable people can disagree about where the lines fall. We document them here in the interest of transparency about how our evaluation operates.

\subsubsection{Human-Human Agreement}
Similar to what was done for Source Quality, we use two domain experts to independently label prompt--response pairs against the binary neutrality verdict, with a senior expert adjudicating disagreements. We report agreement separately on the validation set used during AI judge calibration ($n \approx 100$) and on a held-out test set ($n \approx 160$). We note that any prompt--response which does not pass the neutrality evaluation was automatically evaluated for political lean. This resulted in a separate validation set (($n \approx 60$) and held-out test set ($n \approx 100$). Inter-rater agreement is measured using Krippendorff’s $\alpha$, with $95\%$ confidence intervals estimated via item-level bootstrapping ($2K$ resamples). 

\Cref{tab:neut-human} shows that independent human raters achieve strong consensus on neutrality, with inter-rater agreement of $\alpha = 0.72$, exceeding our minimum acceptable threshold of $0.67$. Agreement on political lean, shown in \Cref{tab:lean-human}, is more moderate ($\alpha = 0.65$), falling just short of this threshold. We note, however, that political lean is a three-category label, making agreement inherently more challenging to achieve. We therefore view these results as promising evidence of rubric reliability, while continuing to iterate on the protocol to further improve consistency.

\subsubsection{AI-Human Agreement} 
Again, we use the calibrated AI judge created during our pipeline to annotate the same held-out test set. We measure reliability using precision, recall, and F1 of the AI annotations against the human gold-labels for further analysis. 

\begin{table}[h]
\centering
\caption{Neutrality: Inter-expert agreement (Human--Human). Krippendorff's $\alpha$ with item-bootstrap 95\% CI on both the validation and test set.}
\label{tab:neut-human}
\small
\begin{tabular}{@{}lcc@{}}
\toprule
\textbf{Standard} & \textbf{Validation} & \textbf{Test} \\
& \textbf{$\alpha$ [95\% CI]}& \textbf{$\alpha$ [95\% CI]}\\
\midrule
Neutrality (bin.) & .79 [.66,.90] & .72 [.60, .83] \\
\bottomrule
\end{tabular}
\end{table}

\begin{table}[h]
\centering
\caption{Political Lean: Inter-expert agreement (Human--Human). Krippendorff's $\alpha$ with item-bootstrap 95\% CI on both the validation and test set.}
\label{tab:lean-human}
\small
\begin{tabular}{@{}lcc@{}}
\toprule
\textbf{Standard} & \textbf{Validation} & \textbf{Test} \\
& \textbf{$\alpha$ [95\% CI]}& \textbf{$\alpha$ [95\% CI]}\\
\midrule
Political Lean (categ.) & .68 [.55,.80] & .65 [.51, .78] \\
\bottomrule
\end{tabular}
\end{table}

\begin{table}[h]
\centering
\caption{Neutrality: Precision, recall, and F1 of the automated AI judge against the adjudicated human consensus on the test set.}
\label{tab:neut-judge-class}
\small
\begin{tabular}{@{}lrccc@{}}
\toprule
\textbf{Standard} & \textbf{$n$} & \textbf{Prec.} & \textbf{Rec.} & \textbf{F1} \\
\midrule
Neutrality (bin.)\\
\textit{macro avg} & 159 & .85 & .85 & .85 \\
\textit{weighted avg} & 159 & .86 & .86 & .86 \\
\bottomrule
\end{tabular}
\end{table}

\begin{table}[h]
\centering
\caption{Political Lean: Precision, recall, and F1 of the automated AI judge against the adjudicated human consensus on the test set.}
\label{tab:lean-judge-class}
\small
\begin{tabular}{@{}lrccc@{}}
\toprule
\textbf{Class} & \textbf{$n$} & \textbf{Prec.} & \textbf{Rec.} & \textbf{F1} \\
\midrule
Political Lean (categ.)\\
\textit{macro avg} & 99 & .72 & .76 & .73 \\
\textit{weighted avg} & 99 & .81 & .78 & .79 \\
\bottomrule
\end{tabular}
\end{table}

The automated neutrality judge reaches $F1 = 0.86$ against the adjudicated human consensus on the held-out test set, as seen in \Cref{tab:neut-judge-class}. Per-class scores are balanced with high recall (.86) and high precision (.86), indicating that the judge is strong for both positive and negative examples. Overall the calibrated judge tracks human neutrality verdicts with strong agreement and balanced performance across both classes.

For political lean shown in \Cref{tab:lean-judge-class}, we find that the AI judge has moderate agreement with experts ($F1 = .79$). In general, we found performance is strongest on explicit left and right lean, and only slightly weaker on the none/other category, suggesting that genuinely ambiguous or mixed cases remain the most challenging to classify. Overall, the calibrated judge appears capable of recovering clear partisan lean while preserving uncertainty in gray-area cases, rather than overconfidently forcing ambiguous examples into partisan labels.

\section{Limitations and Future Work}

The validation results presented here cover two of our three editorial dimensions for the news reporting benchmark. Source quality has relatively objective markers and is the most straightforward to encode. Neutrality is more interpretive but admits a workable binary verdict at the prompt--response level. Factual accuracy is structurally more challenging, as it requires additional pipeline components, including an extraction step prior to any verification, and will be reported in subsequent work.

Our expert panels, while diverse and drawn from multiple professional domains, are finite groups whose consensus defines ground truth for calibration purposes. Expert consensus is not the same as objective truth, and the question of whether a different panel of equally qualified experts would produce materially different calibration targets remains open. Expanding the evaluator network to test the stability of consensus labels across different expert compositions is a priority.

Similarly, our current gold-labeled datasets are constructed to produce a single definitive label for each sample, with senior experts adjudicating cases of disagreement among domain expert annotators. While this approach provides clear calibration targets, it also collapses ambiguity that may itself be substantively meaningful. In future work, rather than resolving all non-consensus cases into a single label, we plan to preserve and explicitly annotate ambiguity, enabling evaluation of a model’s ability to recognize subtle, contested, or context-dependent cases. Expanding evaluation in this way would better capture edge cases and allow assessment not only of whether models reach expert conclusions, but also whether they appropriately recognize when certainty is unwarranted.

While the validation framework and judge calibration architecture are designed to be domain-general, the expert network, editorial standards, and topic taxonomy are domain-specific and must be rebuilt for each new area of application. The political content benchmark is a proof of concept; extending it to healthcare, defense, finance, and other high-stakes domains requires building the relevant expert networks and partnerships.

\section{Conclusion}

This paper presents a methodology for distilling expert judgment into automated evaluation systems that operate at scale, grounded in a long research tradition on the nature and value of expertise. The domains where AI evaluation matters most are precisely the domains where decades of research confirms that structured expert reasoning captures patterns, contextual distinctions, and calibrated uncertainty that neither crowdsourced labeling nor unsupervised AI self-assessment can reproduce. Independent evaluation is also structurally necessary. The entities building AI systems cannot credibly evaluate their own products, and the public interest requires evaluation that is visibly grounded in identifiable human expertise.

We validate the methodology on political content, demonstrating that expert-calibrated judges reproduce expert consensus at levels comparable to inter-expert agreement on both source quality and neutrality, and that uncalibrated frontier models cannot close the gap through prompt engineering alone. As AI systems are deployed on an expanding set of consequential topics, the need for this kind of independent, expert-grounded evaluation infrastructure will only grow. The architecture is designed to generalize to any domain where judgment matters and expertise is scarce, and building out that broader infrastructure is the work ahead.

\section*{References}
\small
\begin{list}{}{\setlength{\leftmargin}{1.5em}\setlength{\itemindent}{-1.5em}\setlength{\itemsep}{2pt}\setlength{\parsep}{0pt}}

\item Chase, W.\ G., \& Simon, H.\ A.\ (1973). Perception in chess. \textit{Cognitive Psychology}, 4(1), 55--81.

\item Chi, M.\ T.\ H., Glaser, R., \& Farr, M.\ J.\ (Eds.). (1988). \textit{The Nature of Expertise}. Lawrence Erlbaum Associates.

\item Dawes, R.\ M., Faust, D., \& Meehl, P.\ E.\ (1989). Clinical versus actuarial judgment. \textit{Science}, 243(4899), 1668--1674.

\item Dreyfus, H.\ L., \& Dreyfus, S.\ E.\ (1986). \textit{Mind Over Machine: The Power of Human Intuition and Expertise in the Era of the Computer}. Free Press.

\item Ericsson, K.\ A., Krampe, R.\ T., \& Tesch-R\"{o}mer, C.\ (1993). The role of deliberate practice in the acquisition of expert performance. \textit{Psychological Review}, 100(3), 363--406.

\item Grove, W.\ M., Zald, D.\ H., Lebow, B.\ S., Snitz, B.\ E., \& Nelson, C.\ (2000). Clinical versus mechanical prediction: A meta-analysis. \textit{Psychological Assessment}, 12(1), 19--30.

\item Hayes, A.\ F., \& Krippendorff, K.\ (2007). Answering the call for a standard reliability measure for coding data. \textit{Communication Methods and Measures}, 1(1), 77--89.

\item Kahneman, D., \& Klein, G.\ (2009). Conditions for intuitive expertise: A failure to disagree. \textit{American Psychologist}, 64(6), 515--526.

\item Klein, G.\ (1998). Sources of Power: How People Make Decisions. \textit{MIT Press}.

\item Krippendorff, K.\ (2019). \textit{Content Analysis: An Introduction to Its Methodology} (4th ed.). SAGE Publications.

\item Meehl, P.\ E.\ (1954). \textit{Clinical Versus Statistical Prediction: A Theoretical Analysis and a Review of the Evidence}. University of Minnesota Press.

\item Shanteau, J.\ (1992). Competence in experts: The role of task characteristics. \textit{Organizational Behavior and Human Decision Processes}, 53(2), 252--266.

\item Tetlock, P.\ E.\ (2005). \textit{Expert Political Judgment: How Good Is It? How Can We Know?} Princeton University Press.

\item Tetlock, P.\ E., \& Gardner, D.\ (2015). \textit{Superforecasting: The Art and Science of Prediction}. Crown.

\end{list}

\section*{AI Acknowledgment}

During the preparation of this work, the authors used Claude Opus 4.7 to help summarize material and improve the readability of the manuscript. After using this tool, the authors reviewed and edited the content as needed and take full responsibility for the content of the publication.

\end{document}